\documentclass[runningheads]{llncs}

\usepackage{eccv}

\usepackage{eccvabbrv}

\usepackage{graphicx}
\usepackage{booktabs}
\usepackage{wrapfig}

\usepackage[accsupp]{axessibility}  

\usepackage{hyperref}

\usepackage{orcidlink}

\begin{document}

\title{EgoEverything: A Benchmark for Human Behavior–Inspired Long-Context Egocentric Video Understanding in AR Environment} 

\titlerunning{EgoEverything}

\author{
Qiance Tang\inst{1}\textsuperscript{$\dagger$}\orcidlink{0009-0006-7804-5454} \and
Ziqi Wang\inst{1}\textsuperscript{$\dagger$}\orcidlink{0009-0003-1579-4990} \and
Jieyu Lin\inst{2}\orcidlink{0000-0002-6875-658X} \and
Ziyun Li\inst{2}\orcidlink{0000-0001-6070-6310} \and \\
Barbara De Salvo\inst{2}\orcidlink{0009-0008-0071-2741} \and
Sai Qian Zhang\inst{1}\textsuperscript{$\ddagger$}\orcidlink{0000-0002-4815-9235}
}
\authorrunning{Q. Tang, et al.}

\institute{ New York University, USA\\  \and Meta Reality Labs, USA\\[1mm] \textsuperscript{$\dagger$} Equal contribution. \quad \\[1mm] }

\maketitle

\begin{abstract}
Long-context egocentric video understanding has recently attracted significant research attention, with augmented reality (AR) highlighted as one of its most important application domains. Nevertheless, the task remains highly challenging due to the need for reasoning over extended temporal contexts and diverse, unstructured activities. Although several benchmarks exist, most egocentric datasets rely on human-worn cameras and focus mainly on visual content, with limited consideration of underlying user behavior when forming video-related queries. EgoEverything is a benchmark that uses real gaze traces as a weak attention prior, rather than as a direct proxy for user intention, when generating questions. It comprises over 5,000 multiple-choice question-answer pairs, spanning more than 100 hours of video. By integrating measured gaze traces with a lightweight spatial sampling prior, it more faithfully captures natural AR-style querying behavior and offers a realistic evaluation setting for long-context egocentric video understanding in AR. We release our dataset at \href{https://sai-lab-nyu.github.io/EgoEverything/}{\texttt{https://sai-lab-nyu.github.io/EgoEverything/}}. 
\end{abstract}

\begin{wrapfigure}{r}{0.6\textwidth}
  \vspace{-29pt}
  \centering
  \includegraphics[width=0.6\columnwidth]{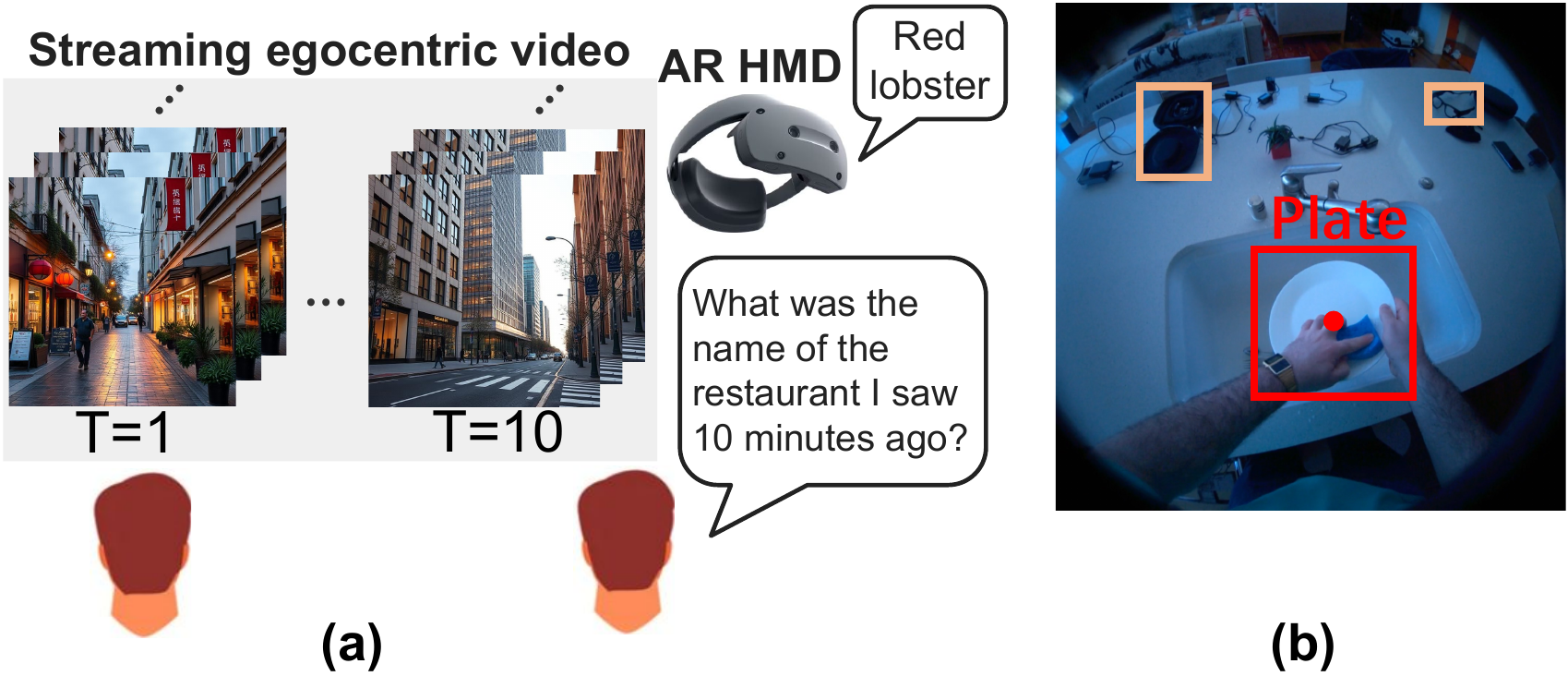}
  \caption{(a) Real-life AR LEU scenario. (b) User attention variation.}
  \label{fig:motivation-examples}
\vspace{-20pt}
\end{wrapfigure}
\section{Introduction}
\label{sec:intro}
Augmented Reality (AR) is emerging not only as a novel user interface but also as a machine learning (ML) platform that integrates embodied experiences such as sensing, perception, memory, and speech. By aligning digital and physical realms, AR enables real-time information extraction and contextual decision-making, transforming domains including healthcare~\cite{gerup2020augmented,chengoden2023metaverse}, education~\cite{westin2022inclusive,al2023analyzing}, and industry~\cite{chidsin2021ar,jo2021virtual}.

Beyond immersive applications, AR devices generate continuous multimodal data streams that are invaluable for ML research. Equipped with cameras, eye-tracking, hand-tracking, and motion sensors, they capture high-bandwidth visual, auditory, and behavioral signals in real time. These heterogeneous streams pose significant challenges but also offer new opportunities for long-context ML models to capture correlations between signals, user attention, and the surrounding environment across extended timescales, enabling more effective assistance for AR users in everyday scenarios.

As illustrated in Figure~\ref{fig:motivation-examples} (a), consider an AR head-mounted display (HMD) used for driving navigation. The device continuously encodes multimodal inputs through ML models. When the user briefly glances at a restaurant, this gaze event must be accurately linked to the corresponding visual features. Later, answering a query such as “What was the name of the restaurant we passed 10 minutes ago?” requires an episodic memory module that can index temporal embeddings, retrieve the relevant instance, and integrate it with a machine learning model (e.g., Vision-Language Model (VLM)) for analysis.
This paradigm transforms AR from a passive data collector into an intelligent personal assistant powered by long-context ML. In everyday life, AR could enable superhuman memory, helping users retrieve details such as where they left their keys. 
In education, students could revisit past demonstrations or experiments to reinforce learning. In healthcare, surgeons might query the system to recall the exact moment when a critical anatomical landmark appeared during a procedure.
Building toward this vision, long-context egocentric video understanding (LEU) has become an increasingly active area of research in the ML community. Long egocentric recordings capture extended daily activities and interactions, requiring models to reason over temporal dependencies and multimodal signals spanning minutes or hours. To evaluate progress, a growing set of benchmarks has been introduced~\cite{perrett2025hdepichighlydetailedegocentricvideo,chandrasegaran2024hourvideo1hourvideolanguageunderstanding,zhou2025x, wu2024longvideobench}, each designed to test how well models can recall, integrate, and reason over such extended sequences. 
While these benchmarks represent important advances, they largely emphasize generic video-based queries and fall short of capturing the~\textbf{human-centric} and~\textbf{attention-guided} nature of real AR usage, where questions are often grounded in what the user was attending to at a given moment. These limitations are summarized as follows:

\noindent\textbf{Questions Not Reflecting Human Attention:} Existing benchmarks rarely consider user attention when designing queries, creating a mismatch with real-world usage. In practice, people tend to ask about objects or events they have looked at, at least partially, or objects near where their attention was directed. Current datasets instead emphasize generic questions about visual details or scene overviews, and fail to reflect human inquiry patterns.

\noindent\textbf{Questions Not Framed in Natural Language:} Prior benchmarks often rely on rigid, template-based question generation that does not align with authentic human questioning. For example, prompts such as “Is the light off in the video?” frequently appear, but they are uncommon in daily use. In contrast, real users are more likely to ask context-specific, attention-driven questions such as “Did I forget to turn off the lights?”

\noindent\textbf{Questions Not Aligned with the Moment of Interaction:} Most benchmarks restrict questioning to occur before or after a clip has ended. However, users typically pose questions \textbf{during} ongoing interactions, requiring real-time reasoning over partially observed streams.


To address these limitations, we present~\textit{EgoEverything}, a benchmark for LEU that {approximates realistic AR-assistant querying scenarios} with real gaze traces and natural language questions. Collecting egocentric video in realistic AR scenarios and manually authoring multiple-choice questions is both labor- and time-intensive. Annotators must repeatedly review videos, verify fine-grained details, craft challenging queries, and refine phrasing to approximate natural user language. As a result, many prior works resort to template-based question generation~\cite{perrett2025hdepichighlydetailedegocentricvideo,xiao2021next, li2023intentqa}, which lowers cost and improves consistency but fails to capture how AR users actually ask questions. In contrast, EgoEverything is constructed through a VQA generation pipeline that leverages multiple AI agents to produce questions aligned with authentic human questioning patterns. We further introduce {a gaze-aware target sampling strategy that uses measured gaze only as a weak attention prior,} enabling the benchmark to include both attention-driven queries and detail-oriented ones outside the user’s focus. This design raises task difficulty while more closely matching real-world AR query behavior. Finally, we incorporate~\textbf{comprehensive human review} to enhance question quality and reliability. Specifically, our contributions are:

\begin{itemize}
\item {We introduce EgoEverything, a gaze-aware LEU benchmark for AR-assistant scenarios, where questions are natural, context-specific, and asked at realistic interaction times.}
\item {We propose a reusable VQA generation pipeline combining gaze-oriented target sampling, multi-agent question synthesis and validation, rule-based filtering, blind filtering, and human review.}
\item {We provide over 5,000 multiple-choice question--answer pairs across more than 100 hours of egocentric video, and show that current VLMs remain far below human performance under attention-aware and long-recall settings.}
\end{itemize}

\section{Background and Related Work}
\label{sec:formatting}

\textbf{Augmented Reality} (AR) devices are rapidly maturing into always-on, wearable interfaces that bridge virtual content with the physical world through lightweight headsets (e.g., Meta Aria \cite{engel2023project}). Modern units typically include a front-facing high-resolution camera (1408 × 1408 on Meta Aria glasses) and support natural interaction via hand gestures and human gaze. Crucially, today’s headsets can track gaze reliably in real time \cite{hou2024unveiling}, providing a solid engineering basis for attention modeling and gaze-conditioned interaction.

Unlike phones or desktops, AR headsets are designed and expected to be worn for extended periods and operate under tight battery, thermal, and memory budgets \cite{engel2023project,goesele2025imagingalldaywearablesmart}. On these devices, compute is embedded (CPU/GPU), but RAM is limited and often shared between CPU and GPU (e.g., on Meta Quest 3  $\approx$ 8 GB \cite{quest_3}), constraining model size and throughput. Meanwhile, a practical AR assistant must continuously process long video streams while the user goes about daily activities. These constraints motivate smaller models and more efficient algorithms, especially streaming, memory-aware methods that compress and sample long-duration visual inputs smartly. With such methods, robust perception, reasoning, and dialogue can run on-device within unified memory limits.


\subsection{Spatial Dynamics of Human Attention}
\label{sec:bg:user-attn}
Human perception is inherently selective, as the brain cannot process the entire visual field with equal fidelity. Prior research has described the attention field as a “spotlight”~\cite{Eriksen1986,Posner1980}, often approximated by gaze location. Attention strength typically decays spatially, commonly approximated by a 2D Gaussian~\cite{Ioannides2010}, resulting in high fidelity at the focus point that gradually diminishes with distance~\cite{desimone1995neural,carrasco2011visual}.

This attentional pattern strongly influences the types of questions an AR user is likely to ask in real scenarios. As illustrated in Figure~\ref{fig:motivation-examples} (b), a user may focus on a cup while walking in the kitchen. Since attention is concentrated on the cup, the user is more likely to later issue a query about this object or its immediate context (e.g., {``Did I leave the cup on the counter?''}). In contrast, nearby items, shown with lighter bounding boxes in Figure~\ref{fig:motivation-examples} (b), receive weaker attention and are therefore less likely to become the subject of subsequent queries. Similar findings have been reported in cognitive psychology and vision science, where gaze serves as a reliable predictor of future memory recall and questioning behavior~\cite{yarbus2013eye,land2001ways}.

In the field of Psychology and Neuroscience, human attention has historically and empirically been described as a field that decays from a focal point in the visual field. Past research has described metaphorically the attention field as a spotlight~\cite{Eriksen1986, Posner1980}, depicted with a gradient model, and sometimes a Zoom-lens Model \cite{Eriksen1986}. This research all pointed to a 2D Gaussian modeling for the human attention field \cite{Ioannides2010}. 

Modern AR headsets provide reliable, real-time gaze tracking, which makes it possible to instantiate this 2D Gaussian prior directly on the device. This allows perception and assistance to be aligned with what the user is actually attending to during everyday, long duration use. The gaze-centric attention representation is also computationally friendly on embedded AR hardware: it enables foveated sampling, gaze-guided segmentation, and streaming compression of long visual sequences. These features are key tactics when CPU/GPU share limited RAM and power budgets on wearable devices.

\subsection{Vision Language Models}
\label{sec:bg:vlm}
Contemporary VLMs \cite{geminiteam2024gemini15unlockingmultimodal, openai2024gpt4technicalreport,zhang2025videollama, deitke2024molmopixmoopenweights, bai2025qwen25vltechnicalreport} extend language-guided foundation models to additional modalities and demonstrate strong, general capabilities. Trained at scale with spatial data \cite{chen2024spatialvlmendowingvisionlanguagemodels}, they achieve high accuracy in spatial understanding and encode rich human preferences and priors. They support perception, reasoning, instruction following, and dialogue, enabling AR assistants that ground semantics in what the user sees and points to. 

\subsection{Long-context Egocentric Video Understanding}
\label{sec:bg:long-context-video-understanding}
Long-context egocentric video understanding has recently gained significant attention in the machine learning community, particularly in extended first-person recordings that capture daily activities and interactions. Such representations enable AR systems to support timely, context-aware assistance by recalling and reasoning over past events in real-world environments.
Recent advances in long-context information representation increasingly emphasize structured approaches~\cite{yang2025memory,wang2023lifelongmemory,yang2025streammem,arnab2021unified,baradel2018object,brendel2011learning,cong2021spatial}. In the egocentric vision domain, studies have explored structured video representations by grouping video segments into activity threads~\cite{price2022unweavenet,fan2024videoagent,yang2025memory} or constructing egocentric scene graphs to model object–user relationships~\cite{goletto2024amego, rodin2024action,huang2025building}. The stored memory entries can later be queried by the user, and these entries are then provided as input to a machine learning model (e.g., VLM) to generate an answer. 

Alongside these advances, numerous benchmarks have been introduced to evaluate long-context egocentric video understanding~\cite{perrett2025hdepichighlydetailedegocentricvideo,chandrasegaran2024hourvideo1hourvideolanguageunderstanding,zhou2025x, wu2024longvideobench, mangalam2023egoschema, xiao2021next, li2023intentqa}. However, these datasets primarily emphasize generic video-based questions or maximum video duration, and overlook the human-centric nature of real AR usage. They also restrict questioning to occur only before or after a clip or fixed segment has concluded. In practice, users tend to ask questions anchored to where their attention was directed, often indicated by gaze during recording, yet existing benchmarks fail to capture this critical dimension. Consequently, they fall short of simulating realistic AR scenarios in which attentional focus fundamentally shapes memory retrieval and contextual reasoning. Empirical evidence further supports this view, as incorporating gaze has been shown to significantly improve grounding in egocentric retrieval and natural language query (NLQ) tasks~\cite{lin2025gazenlqego4dnatural}.

{Several concurrent gaze-aware benchmarks are complementary to EgoEverything. EgoGazeVQA evaluates egocentric intent understanding with explicit gaze guidance and studies textual gaze prompts, visual marks, and saliency-map prompting~\cite{peng2025eyemllmbenchmarkingegocentric}. GazeVQA focuses on multiview eye-gaze task-oriented collaboration in an industrial assembly/disassembly setting~\cite{ilaslan-etal-2023-gazevqa}. StreamGaze studies gaze-guided past, present, and proactive reasoning in streaming videos~\cite{lee2026streamgazegazeguidedtemporalreasoning}. In contrast, EgoEverything uses gaze offline during target sampling, rather than exposing gaze as part of the final query. This design keeps the questions natural and context-specific while jointly satisfying three AR-oriented criteria: reflecting user attention, preserving user-style language, and aligning questions with the moment of interaction.}

\begin{figure}[ht]
  \centering
  \includegraphics[width=0.7\columnwidth]{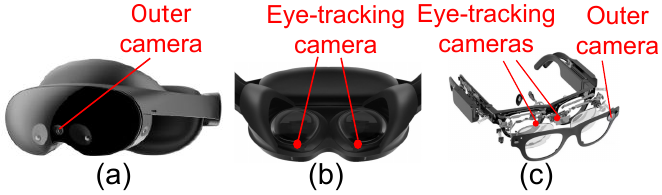}
  \caption{(a) Front and (b) inner views of the Meta Quest Pro headset, which functions as both an AR and VR (Virtual Reality) device. (c) Meta Aria glasses equipped with multiple cameras.}
  \label{fig:ar-device}
  \vspace{-10pt}
\end{figure}

\subsection{AR System}

Figure~\ref{fig:ar-device} illustrates typical AR device hardware configurations. These systems feature multiple front- and side-facing cameras that capture the user's field of view and gaze position. Outward-facing cameras generate high-resolution imagery (e.g., $1408 \times 1408$ on the Meta Aria glasses~\cite{meta_aria}), while inward-facing cameras capture lower-resolution monochrome images of the eyes. Combined, these sensing mechanisms enable gaze tracking~\cite{quest_3,hololens,quest_pro}, typically through either analytical methods such as pupil–corneal reflection modeling or machine learning approaches~\cite{liu2025fovealnet}. These modalities provide critical signals of user attention and interaction, making them foundational for many AR applications.

\section{Data Collection Procedure}

\label{sec:data-collection}
\begin{figure*}[ht]
    \centering
    \includegraphics[width=1\linewidth]{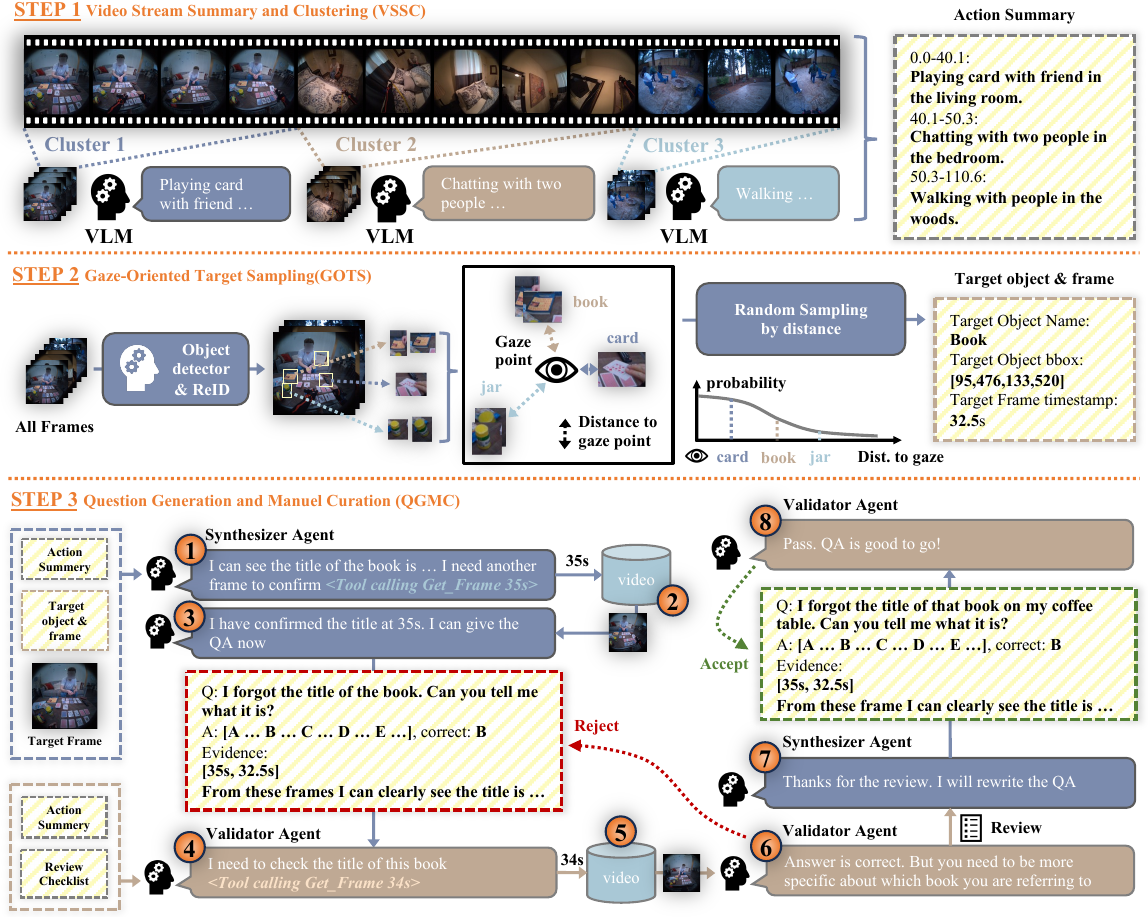}
    \caption{Data generation pipeline for EgoEverything. Step 1, Video Stream Summary and Clustering (VSSC), combines clustering, summary generation, and manual inspection to obtain high-quality video descriptions. Step 2, Gaze-Oriented Target Sampling, detects objects in each frame, computes their distance to gaze, applies ReID to group object instances, and samples target objects by distance. Step 3, Question Generation and Manual Curation, uses a self-feedback Synthesizer Agent and Validator Agent to generate and curate MCQs for each target object.
    }
    \label{fig:pipeline}
    \vspace{0pt}
\end{figure*}
\subsection{Overview}
\label{sec:overview}

The collection process of EgoEverything consists of three steps: (1)~\textit{Video Stream Summary and Clustering} (VSSC), (2)~\textit{Gaze-Oriented Target Sampling} (GOTS), (3)~\textit{Question Generation and Manual Curation} (QGMC). During VSSC, the VLM is prompted with an egocentric video clip to generate a summary of the user’s action over the given time span, as shown in Step 1 of Figure~\ref{fig:pipeline}. Manual inspection is then applied to remove errors in the generated action summaries. Based on the temporal scope of each summary, the egocentric video stream can be categorized accordingly. During GOTS, highlighted in Step 2 in Figure~\ref{fig:pipeline}, each clustered video tile is examined to sample the objects appearing within it. Sampling is guided by the~\textit{Perception Sampler} (PS), which adaptively adjusts its statistical distribution according to the gaze location in the current frame. The selected target objects are then passed to the subsequent stage for question generation. For QGMC, illustrated in Step 3 of Figure~\ref{fig:pipeline}, we deployed two VLM agents to handle question synthesis and validation. The Synthesizer Agent (SA) creates multiple-choice questions (MCQs) centered on the selected target object, whereas the Validator Agent (VA) examines these questions and delivers feedback. Following VQA generation, the dataset undergoes additional refinement through over \textbf{400} hours of human review combined with rule-based screening.




\subsection{Video Stream Summary and Clustering}
\label{sec:VSSC}

EgoEverything is built based on the real-trace egocentric video dataset that included the real gaze tracking trace, including the AriaEveryday Activities (AEA) dataset~\cite{lv2024aria} and Nymeria dataset~\cite{ma2024nymeria}. The AEA dataset includes 143 daily activity clips across 5 indoor locations, totaling $\sim 7.3$ hours, while the {source} Nymeria dataset has $\sim 300$ hours of videos from $\sim 50$ locations. {EgoEverything samples from these sources and contains more than 100 hours of video in total.} Both datasets provide egocentric videos and user gaze points recorded by smart glasses.

While the Nymeria dataset provides rich, time-aligned narration text describing major events and interacted objects, the AEA dataset lacks such annotations. To address this gap, we cluster consecutive frames within each video clip and generate action summaries, following Step 1 of Figure~\ref{fig:pipeline}. 
Specifically, we first extract frame-level visual features using a pretrained ResNet-50~\cite{he2016deep}, and then apply the k-means algorithm to group the frames into clusters.
{For AEA, which lacks activity-level narration, $K$ controls the temporal granularity of VSSC summaries rather than serving as a semantic label. We evaluated $K \in \{8,12,16\}$ and empirically selected $K=12$ because it produced the best downstream question quality in our pipeline. Larger $K$ values yield finer, more verbose summaries, while smaller values merge distinct actions.} To mitigate label jitter, we apply temporal smoothing by assigning each frame the most common label within a $\pm 5$-frame window. Finally, clips with identical labels are merged into contiguous segments while enforcing a minimum segment duration of two seconds.


\subsection{Gaze-Oriented Target Sampling}
\label{sec:GOTS}
Using the video segments obtained from VSSC, we introduce the GOTS framework, which uses real gaze traces for sampling question targets as described in Section~\ref{sec:bg:user-attn} (Step 2 in Figure~\ref{fig:pipeline}). 

GOTS begins by detecting all objects within each video segment, leveraging a VLM to extract their bounding boxes and labels. Because adjacent frames are often nearly identical, this step generates many redundant detections of the same object over time, which diminishes the diversity of potential questioning targets. To mitigate this, we incorporate a lightweight re-identification (ReID) stage. Specifically, each detected object is cropped and encoded using the visual encoder of a pretrained CLIP model~\cite{radford2021learning}. Detected objects with highly similar CLIP embeddings are then consolidated, ensuring only one representative instance of each object is retained across the sequence.
{
\looseness=1
Subsequently, we measure the Euclidean distance between each object's bounding-box centroid and the corresponding gaze position on a per-frame basis. Using this information, we first randomly sample a Target Frame from the video segment, and then sample a single object from the Target Frame as the Target Object based on the PS $S_{\theta}(\cdot)$. We parameterize $S_\theta(\cdot)$ as a 2D Gaussian distribution in its basic form, expressed as:
\par
}
\begin{equation}
\label{eqn:sampling-frequency}
S_{\theta}(\cdot) \propto \exp\!\left(-\frac{\lVert o-f\rVert^{2}}{2\theta^{2}}\right)
\end{equation}

Here $f$ denotes the measured gaze position, and $S_\theta(\cdot)$ gives the selection probability at object centroid $o$. This probability diminishes as the separation $\|o-f\|$ increases, with $\theta$ modulating the decline rate. {For each sampled Target Frame, we select one Target Object. Repeating this sampling procedure across video segments yields multiple target objects, which are then forwarded to the following question-generation pipeline.}

\subsection{Question Generation and Manual Curation}
\label{sec:QGMC}
\subsubsection{Iterative Question Refinement}
\label{sec:SA_VA}
The Synthesizer Agent (SA) generates questions based on the Target Object, mimicking natural human inquiry patterns. Given the \textit{Target Frame} at time $t_0$, we sample a questioning timestamp $t_q$ from $[t_0+\Delta_{\min},\,T]$, where $T$ is the video end and $\Delta_{\min}=3$ minutes. {We define the recall interval for each MCQ as $\Delta=t_q-t_0$.} This randomization avoids trivial overlap with the Target Frame while ensuring diverse temporal coverage. Prior work in cognitive science~\cite{roediger2006test, carpenter2005application, zacks2007event} shows that varying the delay between stimulus and questioning can improve memory and comprehension, and that humans flexibly recall events across different temporal spans. Uniform sampling provides a practical and behaviorally plausible baseline for determining when to ask questions.

We employ a pretrained VLM as the Synthesizer Agent (SA), fine-tuned to invoke external tools through specific APIs. To reduce computational costs, SA does not process the full video directly. Instead, it interacts with two APIs: GetFrame, which retrieves a high-resolution frame at a specified timestamp for static detail analysis, and GetSegment, which provides a downsampled video clip over a selected time span for verifying dynamic activities. Guided by the system prompt, SA constructs an MCQ about the Target Object, following Step 3 of Figure~\ref{fig:pipeline}, with multiple sub-steps. In sub-step 1, SA analyzes the Action Summary and uses the Target Frame to locate the Target Object. The timestamp of the Target Frame provides contextual information about the activity in the Action Summary, while the associated bounding box for object detection guides SA in identifying the visual features of the Target Object. In sub-step 2, SA first drafts a daily life question about the Target Object, reflecting natural human routines, framed in a natural, human-like style. Then, it identifies the required additional information and iteratively invokes tools and evaluates new evidence until sufficient context is collected to construct an MCQ.

The SA then submits the MCQ with supporting evidence to the Validator Agent (VA) for review (sub-steps 3--4). Similar to the SA, the VA accesses the same Action Summary and may also invoke tools to inspect portions of the video during its evaluation. Unlike the SA, the VA does not receive the Target Object or Target Frame. It verifies factual accuracy, identifies ambiguities, evaluates question clarity, and provides at least one additional piece of evidence to enhance the credibility of the MCQ (sub-step 5). The VA returns feedback to the SA (sub-step 6), which refines and resubmits the MCQ to the VA. If all checks pass, the VA finalizes the MCQ. MCQs failing after two review rounds are discarded.

\subsubsection{Manual Filtering and Labeling}
\label{sec:human-filtering}
After question generation, we obtain high-quality MCQs; however, certain failure modes may still lead to low-quality outputs. The most common case arises when the Target Object in the Target Frame is ambiguous due to factors such as distance, occlusion, inadequate lighting, or viewpoint distortion. In such cases, the object detector may misclassify the Target Object as another item. A second failure mode arises when the Agent misinterprets spatial layouts, resulting in view-dependent or incorrect spatial descriptions. Finally, certain target objects are inherently unsuitable for MCQ generation, producing questions that do not match typical AR user queries.

To mitigate these issues, we first apply rule-based filtering to exclude target objects that are unsuitable for questioning (e.g., walls, ceilings, floors, or the camera wearer’s body parts). We further discard MCQs that violate typical AR user questioning patterns, such as those referencing timestamps or explicitly mentioning the word "video."
Next, we conduct human review. Annotators are presented with each MCQ together with the corresponding video. Without access to the correct answer, they are asked to select one choice from five options. If minor issues are observed, annotators may refine the MCQ; for major flaws, they mark the item as invalid. After the review process, we retain only those MCQs whose pseudo answers are consistent with the annotators’ selections.

Finally, we apply large language model (LLM)-based blind filtering: the LLM is prompted to guess the correct answer without access to the video, and we retain only those MCQs it answers incorrectly. This ensures that the retained questions cannot be solved by simple logical reasoning or textual cues alone, preventing video-free answering in LEU.

\section{Evaluation}

\begin{figure}[t]
 \centering
   \includegraphics[width=0.85\linewidth]{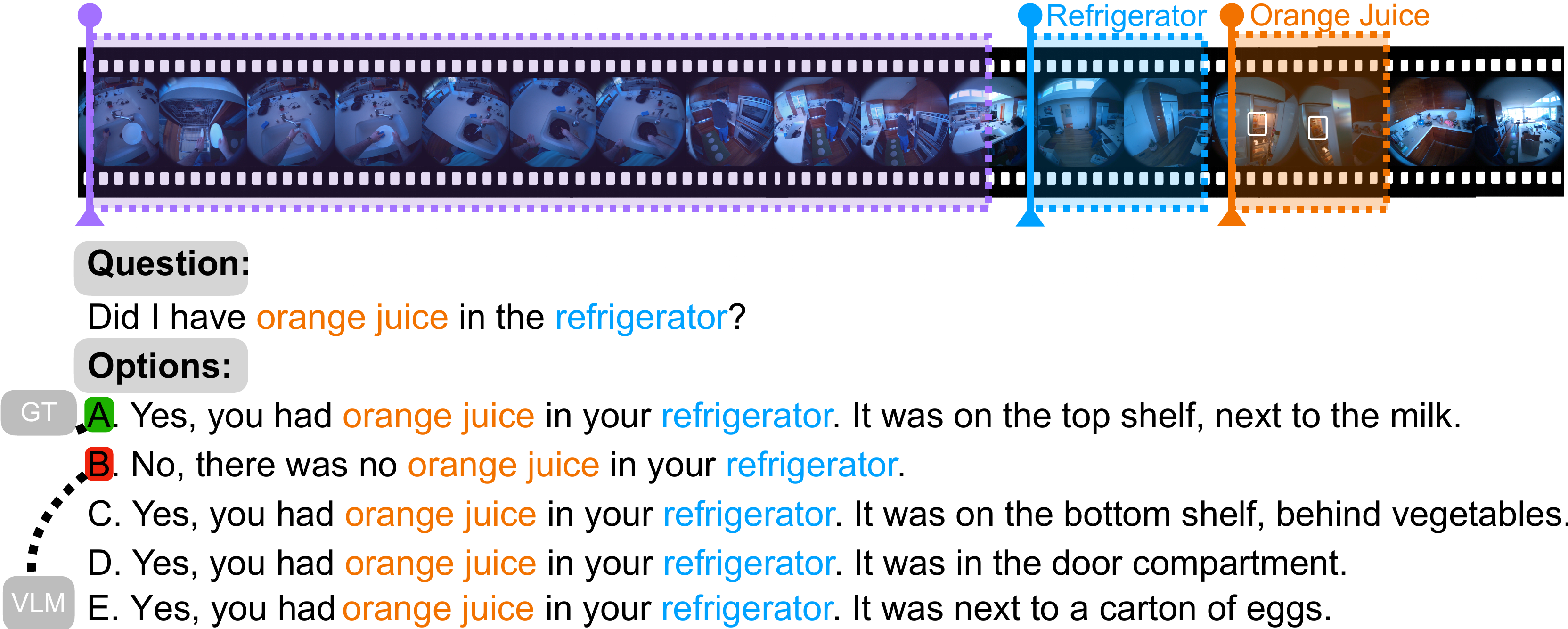}
   \includegraphics[width=0.85\linewidth]{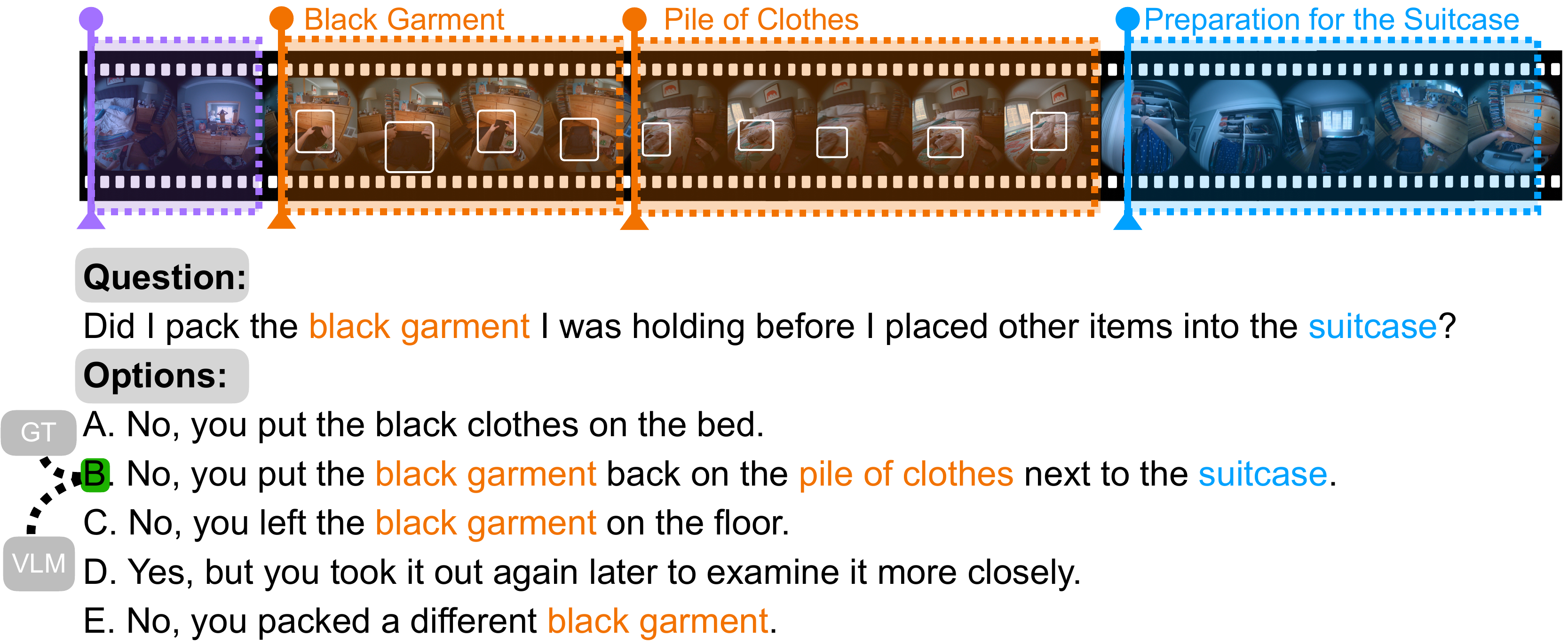}
 \caption{Video frames and keywords in the MCQ are highlighted in different colors. \textcolor{orange}{Orange} indicates the target objects, \textcolor{cyan}{Blue} highlights contextual entities mentioned in the question or answer, and \textcolor{violet}{Purple} denotes regions where the target objects cannot be located. The GT label refers to the ground truth option, and the VLM label refers to the option selected by the model.}
 \label{fig:demonstration}
\end{figure}

EgoEverything is developed on the foundation of real-trace egocentric video datasets enriched with authentic gaze tracking information, incorporating data from the Aria Everyday Activities \cite{lv2024aria} (AEA) and Nymeria \cite{ma2024nymeria} datasets. The AEA dataset comprises 143 egocentric video clips captured at $1408 \times 1408$ resolution (averaging 41.6 MB per minute), spanning five indoor environments with various daily activities across roughly 7.3 hours of footage. Each video is paired with synchronized gaze traces from wearable AR devices, making AEA a compact yet carefully annotated benchmark well-suited for fine-grained studies of attention in everyday tasks. In contrast, the Nymeria source dataset offers a much larger-scale resource, with videos recorded at $2016 \times 2208$ resolution (averaging about 49.3 MB per minute) across nearly 50 diverse indoor and outdoor locations; {EgoEverything uses a filtered subset from Nymeria together with AEA, and the resulting benchmark spans more than 100 hours of video in total.} These videos include real gaze traces along with naturalistic variations in activity, environment, and lighting, providing a rich foundation for training and evaluating models on long-duration and heterogeneous egocentric experiences. Together, AEA and Nymeria complement each other by combining curated activity-focused data with large-scale diverse traces, enabling comprehensive exploration of egocentric understanding. Building on these resources, EgoEverything contains over 5,000 multiple-choice question–answer pairs spanning more than 100 hours of video. Dataset examples are illustrated in Figure~\ref{fig:demonstration}.


In the generation stage, we use the PS in Equation~\ref{eqn:sampling-frequency} with $\theta = 400$ pixels and minimum recall interval $\Delta_{\min}=3$ minutes. {For the $1408 \times 1408$ AEA frames with a $110^\circ$ horizontal field of view, $\theta=400$ pixels corresponds to approximately $31.25^\circ$ in radius. For the $2016 \times 2208$ Nymeria frames, the same radius corresponds to roughly a $40^\circ$ diameter region. Thus, $\theta$ defines a broad central sampling prior around measured gaze rather than a fabricated gaze signal.} For annotation, we developed a web-based labeling tool with 12 trained annotators, who collectively labeled about 21,600 questions over 400 hours. The MCQ adoption rate during annotation was approximately $70\%$, while rule-based and blind filtering yielded acceptance rates of around $50\%$.

We evaluate several recent VLMs on EgoEverything, including Videollama3~\cite{zhang2025videollama}, Gemini~\cite{team2023gemini}, LongVA~\cite{zhang2024longva}, and Llava-Video~\cite{zhang2024video}. Since our MCQs are sampled using a Gaussian-based PS, content near the gaze location tends to be more relevant for solving MCQs than distant information.

To validate this property, we design several preprocessing baselines:~\textit{Gaze Crop} (GC), crops each frame around the gaze fixation location using a square bounding box covering roughly $10\%$ of the original frame size;~\textit{Gaze Mask} (GM), retains the complementary regions outside GC;~\textit{Average Downsampling} (AD), uniformly downsamples each frame to $10\%$ of its original resolution; and ~\textit{Full Resolution} (FR) uses original frames. In addition, we evaluate two recent methods for LEU tasks, AMEGO~\cite{goletto2024amego} and VideoMindPalace (VMP)~\cite{huang2025building}, which construct egocentric scene graphs to capture key object–user relationships while filtering redundant information, achieving strong performance on standard LEU benchmarks such as EgoSchema~\cite{mangalam2023egoschema} and NExT-QA~\cite{xiao2021next}. We apply AMEGO and VMP over Videollama3 and Gemini. Our goal is to examine how these methods perform on this real-life, human attention-driven LEU dataset.

\begin{wrapfigure}[21]{r}{0.50\columnwidth}
  \centering
  \vspace{-35pt}
  \includegraphics[width=\linewidth]{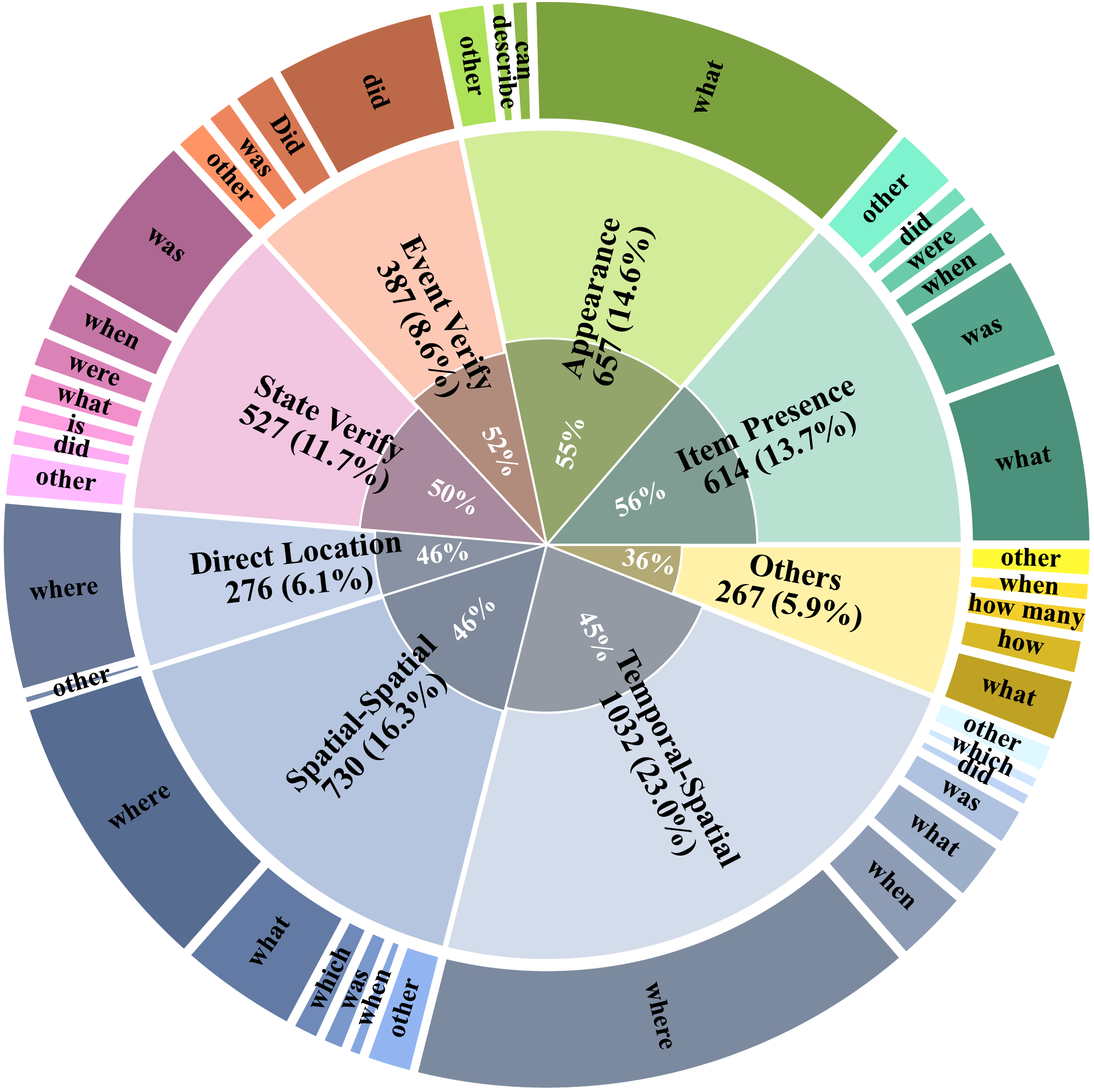}
  \caption{The middle ring shows the proportion of each category and the number of MCQs in it. The outer ring shows the most frequent interrogative words in that category. The inner circle shows the average accuracy of all VLM models from Table~\ref{tab:main} for each category.}
  \label{fig:question_pie}
  \vspace{-6pt}
\end{wrapfigure}

\paragraph{Question Categories}  

We manually classified the generated questions into eight categories:
\textbf{Item Presence}: asking whether an item appears in a given place.
\textbf{Appearance}: asking about the color, shape, or other visual attributes of an item.
\textbf{Event Verification}: asking whether a specific event occurred.
\textbf{State Verification}: asking about the state or condition of an item.
\textbf{Spatial--Spatial}: asking about the relative location between objects.
\textbf{Direct Location}: asking about the absolute location of an item.
\textbf{Temporal--Spatial}: asking where an item is when another event occurs.
\textbf{Others}: miscellaneous questions that do not fit the above categories.
Figure~\ref{fig:question_pie} provides a comprehensive overview of the distribution across question categories.


\begin{figure}[t]
  \centering
  \vspace{-6pt}
  \includegraphics[width=1.0\linewidth]{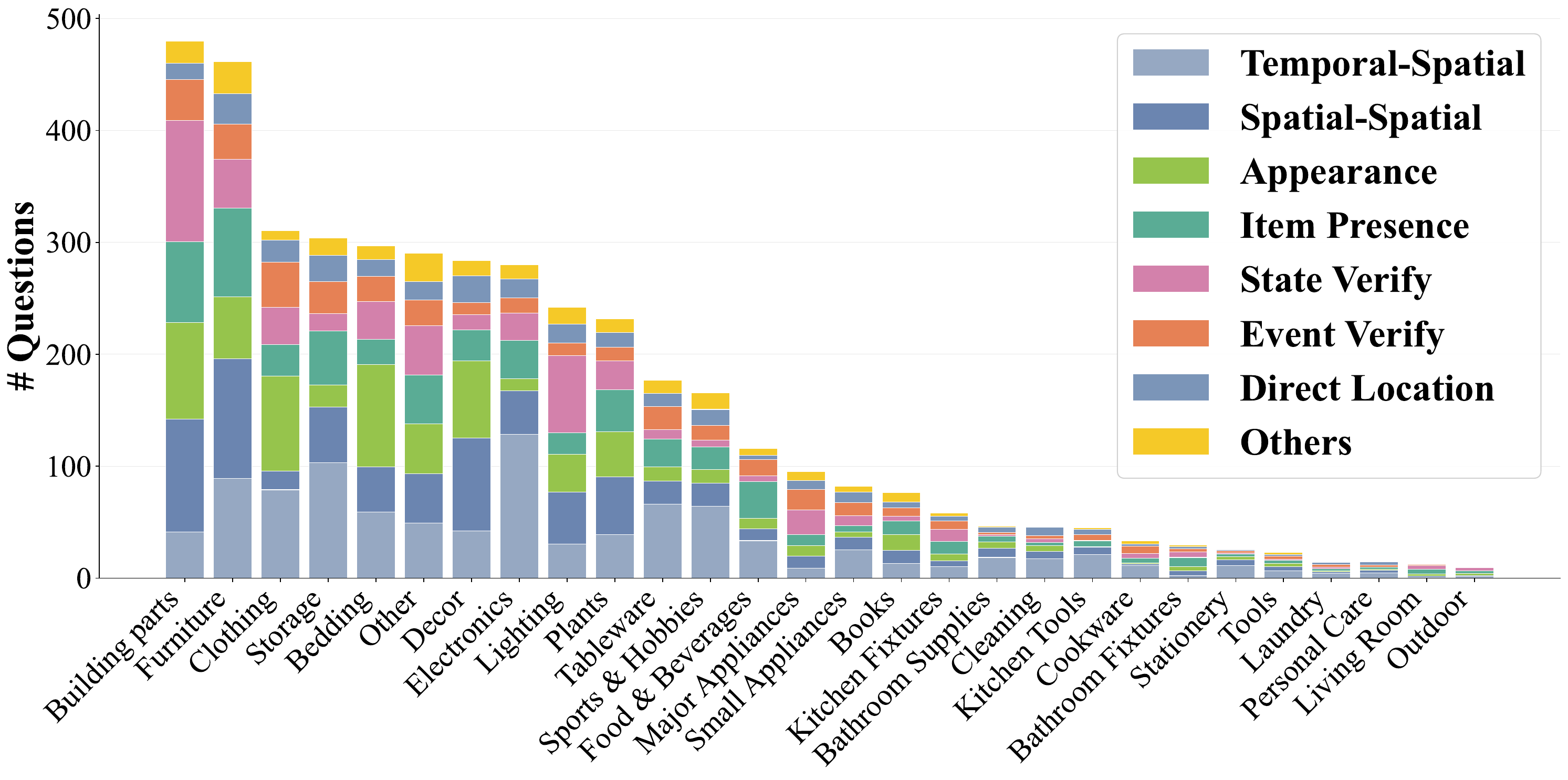}
  \caption{Distribution of target object categories. The x-axis represents object categories, the y-axis counts the number of MCQs generated using the target object in a given category. The length of each colored bar segment indicates the number of questions of that type, revealing the reasoning associated with each target object.}
  \label{fig:key_object_name}
  \vspace{-6pt}
\end{figure}

\paragraph{Target Object Category:} As described in Section~\ref{sec:GOTS}, our GOTS framework selects target objects when generating MCQs. We group selected target objects into 28 categories and show their distribution in Figure~\ref{fig:key_object_name}, where the y-axis denotes the number of associated MCQs and the stacked colors represent the proportions of different question categories within each Target Object category.


\subsection{Accuracy Evaluation on EgoEverything}
\begin{table}[t]
  \centering
  \caption{Performance comparison of different methods across VLMs on EgoEverything. "NA" means not available. The human annotators achieve an average accuracy of 83.5\%.}
  \label{tab:main}
  \resizebox{0.85\linewidth}{!}{%
    \begin{tabular}{lcccccc}
      \toprule
      Model & FR & AD & GC & GM & VMP & AMEGO \\
      \midrule
      Videollama3-7b~\cite{zhang2025videollama} & 49.1 & 46.1 & 42.4 & 35.2 & 20.2 & 19.5 \\
      Videollama3-2b & 46.5 & 44.9 & 40.5 & 34.5 & 21.3 & 19.4 \\
      Gemini 1.5 pro~\cite{team2023gemini} & 63.1 & 58.4 & 52.7 & 37.7 & 33.2 & 18.3 \\
      LongVA~\cite{zhang2024longva} & 34.6 & 31.6 & 28.9 & 22.2 & NA & NA \\
      Llava-Video~\cite{zhang2025videollama} & 42.6 & 36.6 & 32.9 & 26.0 & NA & NA \\
      \bottomrule
    \end{tabular}}
\end{table}

As shown in Table~\ref{tab:main}, among the VLMs, under the full-resolution setting where the entire input frame is provided for processing, Gemini achieves the highest accuracy of $63.1\%$ across the MCQs, while the other models perform worse. Nevertheless, all VLMs remain far behind human performance, as our human annotators reach an average accuracy of $83.5\%$ across 12 participants. This gap highlights the clear performance deficiency of current VLMs on EgoEverything. {We additionally evaluate a saliency-map-based gaze prompting baseline on VideoLLaMA3-7B, following the protocol of EgoGazeVQA~\cite{peng2025eyemllmbenchmarkingegocentric}. This richer gaze input improves accuracy to $51.19\%$, compared with $49.1\%$ under FR, $42.4\%$ under GC, and $35.2\%$ under GM. This result confirms that richer gaze representations can help VLMs, while the remaining gap to human accuracy indicates that EgoEverything is far from solved.}

All input processing methods, including GC, GM, AD, VMP, and AMEGO, show a degradation in MCQ prediction accuracy compared to FR. Among them, GC performs better than GM because the MCQs are generated according to the PS centered at the gaze fixation location, and this method preserves the most critical information. In contrast, GM suffers a substantial performance drop since it excludes these key details regarding human attention specified by the gaze fixation location. Interestingly, AD outperforms GC by preserving not only the gaze-centered visual content but also peripheral information that remains important for answering MCQs. Finally, both VMP and AMEGO achieve the lowest performance among the evaluated methods. This is because neither approach infers directly from raw video. Instead, they convert the video into structured text via object detection and use an LLM for text-only reasoning. Both methods only detect interacted objects and overlook many activity-irrelevant objects in our benchmark.

\begin{wrapfigure}{l}{0.5\textwidth}
  \centering
  \resizebox{\linewidth}{!}{%
    \begin{tabular}{lccc}
      \toprule
      Model & Videollama3-7b & Gemini 1.5 pro & LongVA\\
      \midrule
      Accuracy (\%)   & 22.9 & 35.9 & 21.8  \\
      \bottomrule
    \end{tabular}}
  \captionof{table}{VLM model performance under blind setting.}
  \label{tab:blind}
  \vspace{-15pt}  
\end{wrapfigure}

In addition, to test whether the MCQs within EgoEverything can be solved by solely inspecting the textual information from the question, we conduct a text-only evaluation. This setting is undesirable because it does not allow the model to leverage visual information during processing, and thus may encourage reliance on linguistic shortcuts instead of true multimodal reasoning. {To achieve this, we provide only the question text and answer choices of each MCQ to the VLMs, without any video frames.} As indicated by Table~\ref{tab:blind}, this will greatly degrade the accuracy for Videollama3-7b~\cite{zhang2025videollama}, Gemini~\cite{team2023gemini} and LongVA~\cite{zhang2024longva},  confirming that our questions cannot be answered from text alone.

As described in Section \ref{sec:QGMC}, when generating MCQs we mimic real-life AR scenarios by introducing randomized questioning times $t_q$, whereas in other LEU datasets questioning typically occurs only after a clip or fixed segment has ended. To study how variation in $t_q$ impacts accuracy, we select the $10\%$ of MCQs with the largest recall intervals ($\Delta$) and the $10\%$ with the smallest intervals, and then measure the accuracy of Videollama3-7B over EgoEverything. As shown in Table~\ref{tab:gaze_bbox_recall_layout} (c), the results reveal that accuracy drops markedly as the recall interval increases, from $49\%$ to $33.3\%$. This indicates that current models exhibit substantial performance inconsistency under human-like diverse questioning times, demonstrating that variation in questioning time directly impacts LEU accuracy and thereby validating our randomized questioning-time design.
{VideoLLaMA3 can process videos longer than the maximum recall interval used here, so the degradation more likely comes from redundant visual information, increased distractor activities, and the need to retrieve the correct gaze-grounded episode from a longer temporal stream.}

\paragraph{Impact of Object-Gaze Distance}  

During generation, we sample a target object according to its distance from the gaze point $\lVert o - f \rVert$, as defined in Equation~\ref{eqn:sampling-frequency}. This better simulates human attention and produces questions that align with user focus. To examine the impact of object–gaze distance, we report Videollama3-7b’s accuracy on our MCQs grouped by $\lVert o - f \rVert$, as shown in Table~\ref{tab:gaze_bbox_recall_layout} (a). The results show that target objects located at the periphery of the visual field produce more challenging MCQs than target objects near the center of attention with accuracy decreasing from {$54.8\%$ in the $100$--$200$ pixel bin to $31.4\%$ in the $600$--$700$ pixel bin}. These results indicate that current models struggle with MCQs related to peripheral information. Unlike prior benchmarks, our attention-aware generation produces both challenging peripheral questions and user-focused questions.

\begin{wrapfigure}[22]{r}{0.55\textwidth}
  \centering
  \captionof{table}{Accuracy analysis under three factors. Dist. denotes $\|o-f\|$ (pixel), area denotes bounding box area (pixel$^2$), and recall interval denotes $t_q-t_0$.}

  \resizebox{\linewidth}{!}{%
  \begin{tabular}{cc|cc}
      \toprule
      \multicolumn{2}{c|}{\textbf{(a) Dist. Gaze }} &
      \multicolumn{2}{c}{\textbf{(b) BBox area}} \\
      \midrule
    {Dist. bin (pixel)} & Acc. (\%) & Area (pixel$^2$) & Acc. (\%) \\
    \midrule
    {$0$--$100$} & 54.2 & 100  & 44.3 \\
    {$100$--$200$} & 54.8 & 500  & 44.4 \\
    {$200$--$300$} & 51.8 & 2k   & 44.5 \\
    {$300$--$400$} & 47.2 & 5k   & 44.9 \\
    {$400$--$500$} & 42.6 & 10k  & 49.2 \\
    {$500$--$600$} & 36.1 & 50k  & 54.8 \\
    {$600$--$700$} & 31.4 & 100k & 54.8 \\
      \bottomrule
 \end{tabular}}

  \vspace{10pt}

  \resizebox{0.55\linewidth}{!}{%
  \begin{tabular}{cc}
    \toprule
    \multicolumn{2}{c}{\textbf{(c) Recall interval}} \\
    \midrule
    Percentile & Acc. (\%) \\
    \midrule
    top 10\%    & 33.3 \\
    bottom 10\% & 49.0 \\
    \bottomrule
  \end{tabular}}

  \label{tab:gaze_bbox_recall_layout}
\end{wrapfigure}





\paragraph{Impact of Object Size}

To examine whether VLMs attend to objects of different scales in a comparable manner, we measure bounding box sizes and analyze accuracy as a function of target object area (in pixels$^2$). Area thresholds ranging from $1 \times 10^0$ to $1 \times 10^5$ pixels$^2$ were applied. As shown in Table~\ref{tab:gaze_bbox_recall_layout} (b), accuracy consistently increases with larger bounding box thresholds. Across the evaluated models, performance is biased toward larger objects, while smaller and less salient objects are frequently overlooked.

In summary, these results expose systematic limitations: the evaluated VLMs perform worse on objects that are farther from the gaze point, require longer recall intervals, or appear at smaller scales. As a result, current models still lack the robustness needed for reliable everyday assistance in AR scenarios.

\section{Conclusion}

We introduced EgoEverything, a benchmark for long-context egocentric video understanding that {uses measured gaze traces as a weak attention prior} in AR settings.
EgoEverything leverages {a gaze-aware target sampling strategy grounded in real eye-tracking traces} and a multi-agent VQA generation pipeline, followed by rule-based screening, blind filtering, and extensive human review, to produce over 5,000 high-quality multiple-choice questions spanning more than 100 hours of egocentric video. Our evaluation across several state-of-the-art VLMs reveals a substantial gap to human performance and consistent failures on attention-driven and long-recall questions, highlighting that current models still struggle to reason over long temporal contexts under realistic, user-centric querying patterns. We hope EgoEverything will serve as a practical testbed for robust, attention-aware long-context video understanding in AR assistants.


%
%
\newpage
\bibliographystyle{splncs04}
\bibliography{main}
\end{document}


\clearpage
\setcounter{page}{1}
\title{\small EgoEverything: A Benchmark for Human Behavior–Inspired Long-Context Egocentric Video Understanding in AR Environment\\Supplementary Material}
\maketitle
\appendix

\section{Generation Pipeline: Additional Details}

\paragraph{Prompt Designs}
Prompts for two stages: a video summarizer that identifies current actions and scene and produces a structured summary using VSSC segment boundaries (Sec.~3.2), and an object detector that enumerates visible objects with descriptive names and bounding boxes for consolidation in AGOS (Sec.~3.3).

\begin{figure*}[h]
  \centering
  \begin{subfigure}{0.49\linewidth}
    \centering
    \includegraphics[width=\linewidth]{figures/promt Video_summ_cropped.pdf}
    \caption{Video summarizer prompt, which identifies current actions and scene and enforces a structured output format. Segment metadata are provided by VSSC (Sec.~3.2).}
    \label{fig:appendix_summ}
  \end{subfigure}\hfill
  \begin{subfigure}{0.49\linewidth}
    \centering
    \includegraphics[width=\linewidth]{figures/prompt A.1_cropped.pdf}
    \caption{Object detector prompt. Enumerates visible objects with descriptive names and bounding boxes for subsequent consolidation in AGOS (Sec.~3.3).}
    \label{fig:appendix_objdet}
  \end{subfigure}
  \caption{Prompts used in the generation pipeline. Left: video summarization. Right: initial object detection.}
  \label{fig:appendix_prompts}
\end{figure*}

\begin{figure}[]
    \centering
    \includegraphics[width=0.4\linewidth]{figures/promt Video_summ_cropped.pdf}
    \caption{Prompt for video summarization that identifies the user’s current actions and scene and enforces a structured output. The VLM receives segment boundaries from VSSC Sec.~3.2.}
    \label{fig:Appendix_summ}
\end{figure}

\begin{figure}
    \centering
    \includegraphics[width=0.4\linewidth]{figures/prompt A.1_cropped.pdf}
   \caption{Prompt used for initial object detection in our AGOS pipeline. It instructs the model to identify all clearly visible objects with bounding boxes and descriptive names, which are then passed to the subsequent ReID stage described in Sec.3.2.}
    \label{fig:AppendixA1}
\end{figure}

\begin{figure*}
    \centering
    \includegraphics[width=1\linewidth]{figures/prompt A.2.pdf}
    \caption{We present the prompt used in GA. This prompt specifies the agent’s responsibilities, detailed step-by-step instructions, the required question output format, guidance on deciding whether to invoke tools, and things to avoid. At each step, the information returned to the agent originates from the tools it can call, namely video segment analysis and frame analysis.}
    \label{fig:AppendixA2}
\end{figure*}

\begin{figure*}
    \centering
    \includegraphics[width=1\linewidth]{figures/prompt A.3.pdf}
    \caption{We present the prompt used in RA. This prompt specifies the agent’s responsibilities of Fact Verification and Ambiguity Review, detailed step-by-step instructions, the required review output format, including decision and feedback. At each step, the information sent to the agent originates from the Figure~\ref{fig:AppendixA1} and Figure~\ref{fig:AppendixA2}}
    \label{fig:placeholder}
\end{figure*}